# MailLeak: Obfuscation-Robust Character Extraction Using Transfer Learning

**Wei Wang[1*], Emily Sallenback[2*], Zeyu Ning[1*], Hugues Nelson Iradukunda[1], Wenxi Lu[3], Qingquan Zhang[1,2], Ting Zhu[1]**

[1]University of Maryland Baltimore County, Baltimore, MD 21250, United States.
[2]University of Illinois at Urbana-Champaign, Urbana, IL 61801, United States.
[3]Ohio State University, Columbus, OH 43210, United States.
[*]Co-primary author

## Abstract

The following work presents a new algorithm for character recognition from obfuscated images. The presented method is an example of a potential threat to current postal services. This paper both analyzes the efficiency of the given algorithm and suggests countermeasures to prevent such threats from occurring.

## 1. INTRODUCTION

Physical mail has been a common form of communication for generations. Even with alternative electronic methods, mail is still heavily relied on. Currently, especially in the United States and some European countries, mail is still widely considered to be more reliable than electrical information transfer like emails. Some important information can only be transferred by mail, such as credit cards, bank checks, social security numbers (SSSs), etc.
However, the use of mail is generally not as safe as people think or hope it is. During the process of mailing, a mail can pass through quite a few people, including post offices, transferring drivers, and it may even get lost and returned by some complete strangers. Usually, the information in mail is still considered to be safe if the envelope is not opened, but in the following parts we will show that this is not always the case.
Since the information contained in mail is often very crucial, and a single envelope can pass among multiple different people as it travels, there is a need to protect against malicious man-in-middle attacks. Security patterns are patterns printed on the envelope to distort the information contained in the mail, and it's widely used to hide bank statements, SSNs, and other important information enclosed in envelopes.

The following sections of this paper will focus on using a transfer learning method to conduct the attack on information stored inside an envelope. The method can ignore the effect of some common security patterns and extract the information successfully. It is also extensible, since it can be generalized to different security patterns.

### 1.1 Problem Setting

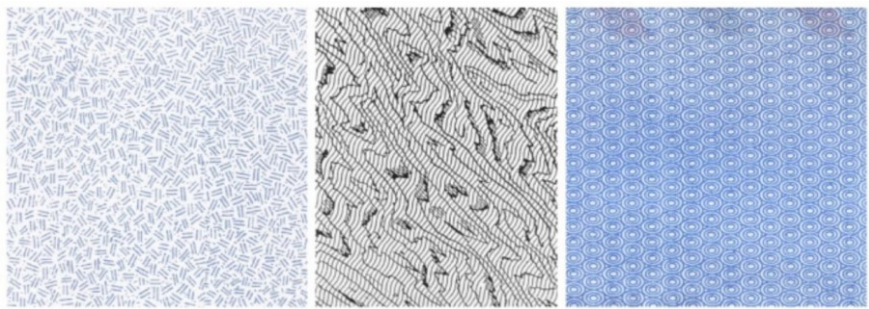

**Figure 1.** Various security patterns used on mails.

Some of the security patterns usually used on envelopes are shown in the following Fig. 1. Although these security patterns generally increase the difficulty of a human reading the contents, the recurring nature of these patterns has made them quite susceptible to attacks based on computer vision. Therefore, this kind of defense is not so reliable when we use machine learning to extract information.

The difficulty with machine learning applications in this context is that the algorithm will not know what specific security pattern is being used. If we directly train a computer vision model solely on examples of one security pattern, the algorithm generally will not work if the mail is using another one. Best stated by Zhang, Sobelman, and He, pertaining to their own study, "The model must be trained before it can be used, which is a limitation of any probabilistic model. The accuracy of the model, therefore, depends on the accuracy of data used to train it." [1], Thus, it is important to train the model on a variety of common security patterns.

Furthermore, OCR can be considered a type of fingerprinting technology because it maps a large number of pixels in an image to a single character, and then from characters to words. Generally speaking, "...fingerprinting technology needs lots of precollection of data." [1] The larger the dataset to test over is, the more accurate the OCR algorithm will be.

If this kind of attack is proved to be real-world applicable, we need a new method of generating the security patterns to resolve the threat.

So generally, the problems and the challenges of this project include:

- Build and test a machine learning-based model to recognize, extract the character information from the semi-transparent picture of the mail;
- Find a way to generalize the model to various security patterns, with relatively low loss on accuracy;
- Design and analyze a security pattern generation procedure to resolve the threat.

## 2.  BACKGROUND

MITM attacks on physical mail are easier to carry out now more than ever. All a person would need is to take a picture of an envelope, ideally while holding it up to a light source, to extract the information inside with an algorithm. Smartphones regularly extract a variety of information from the surrounding environment - so much so that some researchers have suggested companies expand its capabilities to worldwide weather collection, health monitoring, and more[3]. Yet, while smartphone devices can help many, they also pose many security threats. With one click of a photo with a smartphone, a man-in-the-middle can read messages hidden inside envelopes. In an ideal world, secretive messages in mail could be encrypted with cryptographic keys. Then, the security threat relies on the key management, which can be protected through various means[4].

## 3.  RELATED WORK

Recent advances in machine learning[10], big data[45-51] and networking [27, 28, 62-64] have shown the effectiveness in supporting smart applications[21, 52]. However, researchers have shown that these techniques also introduce new security and privacy challenges including defending against new types of attacks[22-26], requiring light-weight authentication and key management system[58-60], proposing novel security mechanism[54, 57, 61] and frame work[53, 55, 56], etc. This work is much inspired by DolphinAtack: Inaudible Voice Commands[5]. Although the methodology is significantly different, the structure and ideas are similar. Our study is also aimed at making a hidden attack which cannot be sensed by victims at all.

The idea of using neural network on OCR is not new; research on this topic has been published as early as 1992[6]. But the real-world application of neural network did not start until the Convolutional Neural Networks (CNNs) are found useful in the research paper by Hinton[7], which is used in our study here. While ANNs (Artificial Neural Networks) are more advanced with more connected nodes, CNNs are more suitable for recognition processes like OCR[8]. Furthermore, the attention mechanism used in our research is originated from the research by Google in 2017[9].

The method presented in this paper is to read the contents of mail uses the context of the mail to help decipher the rest of the contents. This algorithm is similar to that of password guessing algorithms. Yungyu Liu et al., created an efficient password guessing model termed "GENpass"; their model utilizes a vast database of common passwords to help increase the efficiency of their model[10]. Similarly, while the algorithm in this paper does not rely on the context of the paper to read the mail enclosed in an envelope, it does increase the accuracy.

Our approach does not assume that the mail is all typed in a certain font or typed at all. In fact, the tests were done on hand-written documents. Hand-writing introduces another level of complexity since every person has their own unique style. Researchers have shown that how to analyze human gestures through ambient RF signals cited the differences between gestures from person to person as one of the main challenges when it came to efficiency and accuracy of their analysis[11, 18-21]. Similarly, the algorithm in this paper tries to account for human error in the mail.

Although the security of mail is very important in our daily life, the research in this field is relatively scarce. Some remotely-related research or patents[12] mainly focus on the systematic security of mail transfer, but the actual danger lies is ignored. So generally, this is still a relatively new field where very few people have touched. As future work, we also plan to investigate the security and privacy issues in IoT networks[29-34, 42, 65-68], smart grids[35-39, 43-44, 71-79], smart health[40-41, 80], smart agriculture[35], and localization[69-70].

## 4. DESIGN

In order to tackle the problems mentioned, some investigations on the current machine learning algorithms are made. There are already some mature algorithms used for character recognition. Some of the existing algorithms can reach relatively high accuracy even on different lighting conditions, but none of them has discussed the possibility of distortion posed on them. As a result, we applied transfer learning method to solve the first problem mentioned.

### 4.1 Transfer Learning

As a very popular field in machine learning community, transfer learning involves using a large dataset to train an initial model and then transfer it to train for the second phase on a relatively small dataset[13]. This method can save time and effort. It is especially useful when we only have a relatively small database for some specific areas of our interest, since we can tune our model from one which is pre-trained on larger dataset to one which has a relationship to our target domain.

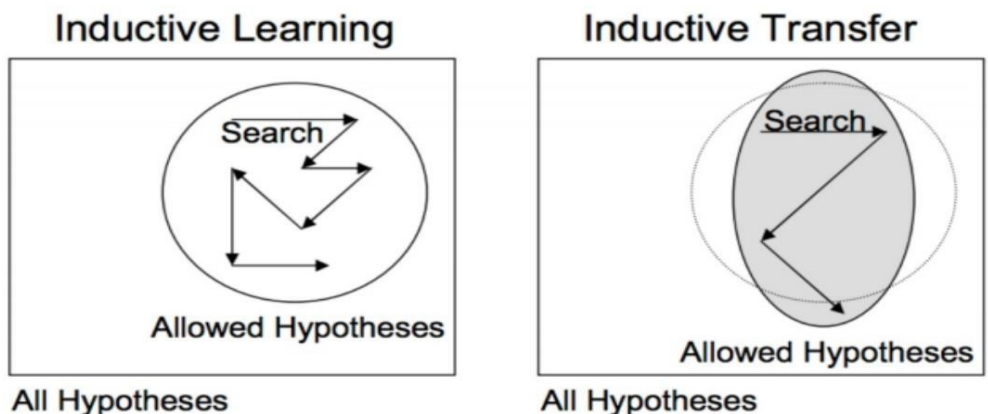

**Figure 2.** Illustration of transfer learning[14].

As shown in the Fig. 2, if we use the obfuscated images to directly train the model, the result might be good for one kind of security pattern, but it's very likely that it cannot be transferred to another one. Thus an open-source model[15] for optical character recognition is used as a base model, this model is first trained on a subset of Synth 90k dataset[16], and we fine-tuned the model via some man-made obfuscated image sets with different security patterns.

## 4.2 Convolutional Neural Networks

Convolutional Neural Networks, known as CNNs are widely used in computer vision and other image-related machine learning tasks. CNNs use learnable filters (convolution kernels) that move along the length and width of the image. These filters will convolve along the length and width of the image (with pre-determined step size) and output the result for the next layer of neural network to process.

## 4.3 Attention Mechanism

The initial model we used also has an attention network, which is widely used in sequential neural networks. Instead of using the full-length of the input vector to predict output, the network will try different combination of weights on different positions. This is especially important for those inputs with sequential input and output, but they are not necessarily linearly corresponding to each other.

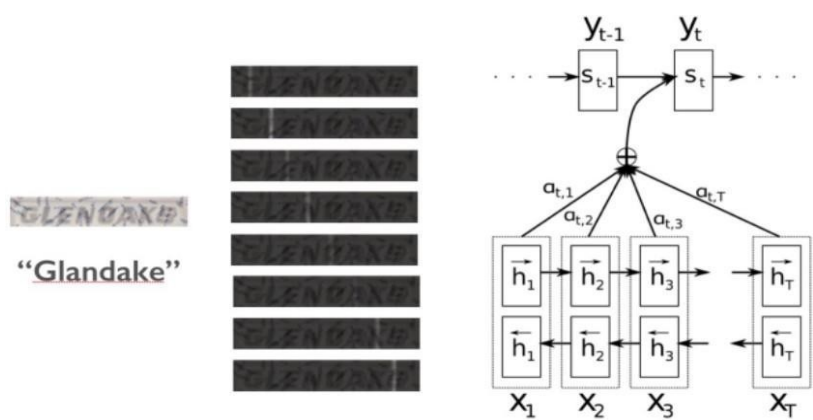

**Figure 3.** Illustration of attention mechanism, using one of the test examples

As shown in Figure 3, In our case, the sequential input from the moving CNN may not correspond directly to the output word in an one-to-one fashion, since the image output may contain quite some blank spaces; therefore, the use of an attention mechanism is important.

## 5. EVALUATION

### 5.1 Network Structure

The first part of this network has 4 CNN layers with 3 batch normalization layers used in between. The kernel size of the CNN layer is 3 x 3, and the step size is 1 pixel per step. Then, the outputs of the CNN layers are fed into a 2-layer attention network, each with 128 units. The LSTM layers in the original model is discarded to improve the speed of training, since the attention in the original model and the spell check model can be as useful as the LSTM layer.

### 5.2 Dataset Generation

We used automatically generated datasets for training and testing purposes along with a real-world test to see if our model can be applied in real-world cases. The training data and the evaluation data are stacked with the security patterns to obscure the characters.

## 5.3 Simulated Test Results

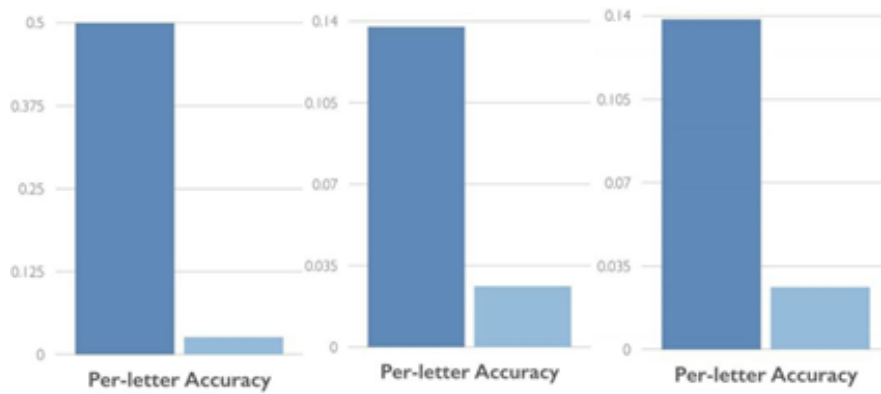

**Figure 4.** Test results for security pattern 1

**Figure 5.** Test results for security pattern 2

**Figure 6.** Test results for security pattern

Figure 4, Figure 5, and Figure 6 show the test results. These results originated from different security patterns, but they all used the same model to show the effect of transfer learning. The model is trained on the first security pattern 1 and tested on all 3.

## 5.4 Spell Check to Improve Accuracy

Since the information in mail is usually words, and our initial approach did not cover the checking of the words' spelling; it's logical to add such a module to improve the accuracy of recognition.

This model is added in two ways:
- With a general package to check the spelling of the word;
- When the attacker has some basic knowledge of the surrounding environment, the attacker can build a dictionary to guess the most possible word under that prior knowledge. In our case, this model is made by a collection of known word labels.

Figure 7 and Figure 8 have shown different cases under the first methods mentioned above:

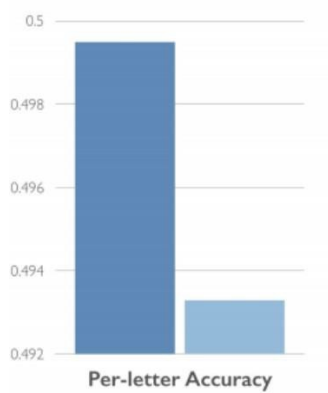
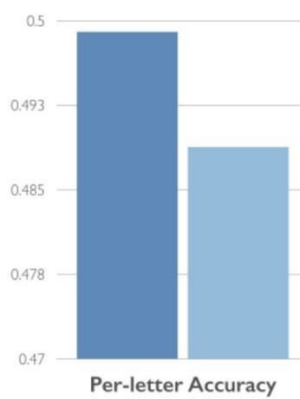

**Figure 7.** Test results with/without general grammar check

**Figure 8.** Test results with/without domain-specific grammar check

## 5.5 Real-world Test

Besides the simulated test shown above, we have also tested the model using some real-world examples. The real-world examples are photos taken directly from semi-transparent envelopes. This model has shown strong results with correct guesses, indicating that this is an effective method in attacking mails in real world, and the current security patterns are not effective enough.

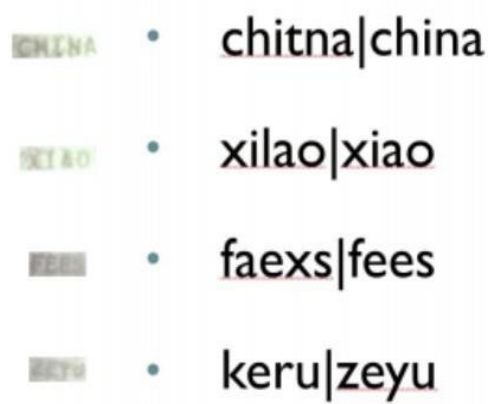

**Figure 9.** Test results with/without domain-specific grammar check

## 6. COUNTERMEASURES

### 6.1 Context-Related Security Pattern

Since the reason where this kind of attack is possible origins from the periodical nature of the current security pattern, we develop a countermeasure to tackle this problem. If the security pattern can change according to the information on the mail, ideally, the mail can be effectively obfuscated even to an extent that all the blank spaces are filled with black blocks. Thus the attacker will definitely have no way to tell the difference between the background and characters, since they all look similar.

This countermeasure is different from simply printing black envelopes. Although printing the envelope black can sometimes be considered as an obfuscation method since it will significantly decrease the transparency of the envelope paper, the contrast ratio between the character and the background does not change under this case.

Our countermeasure against the attack is basically a context-related shader applied on the envelope. Using this kind of countermeasure has some realistic reasons; since the people during the mailing process cannot really know what's in the mail, the mechanism to generate the security patterns would have to be automatic, without any human involvement.

A light-sensitive paper would be an effective solution. When mail is sealed and about to be sent, light travels through the envelope. The light-sensitive ink on the envelope will be opaque on the part which are lighted up. This mechanism, combined with a normal security pattern, is strong enough to obfuscate the attacker; even when the paper in envelope moves, and the shader does not exactly match the position.

Due to the limited time and budget, we have simulated the result of this mechanism by creating a shader layer on the original test images. The result is shown as follow:

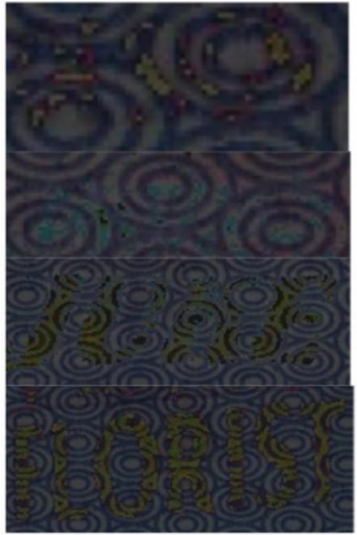

**Figure 10.** Some countermeasure shader altered samples

## 7. ISSUES DURING IMPLEMENTATION

### 7.1 Overfitting

Overfitting is always a big problem in transfer learning. In this research, we have tried several different sets of hyperparameters to make the model less prone to overfitting. We have chosen the iteration number to be around 40,000 to optimize the result. Some other hyperparameters are also carefully tuned.

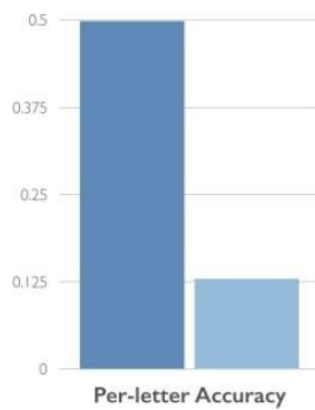 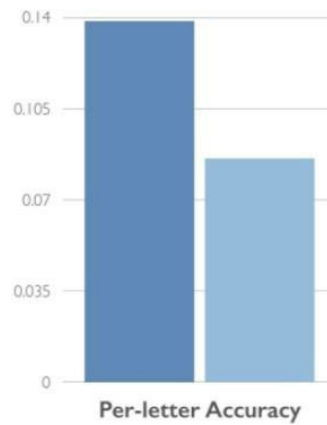

**Figure 11.** Test results for security pattern 1, using countermeasure shader (on the right)

**Figure 12.** Test results for security pattern 3, using countermeasure shader (on the right)

### 7.2 Discussion of the Grammar Model

Whether or not the grammar model has increased the success rate of attack is slightly confusing. Generally speaking, when the single-letter accuracy is higher than 30%, using the grammar model to check the spelling would greatly improve the accuracy, especially when the attack has a knowledge of the surrounding environment and created a related knowledge base. But if the attack has no prior knowledge of the surrounding environment or the accuracy is strikingly low for some cases, the grammar model might take the results further away from the reality.

While mail may contain misspellings and grammatical errors, we found that the grammar model included in the algorithm does not increase the accuracy significantly unless there is some detailed knowledge of the surrounding environment.

# 8. CONCLUSION AND FUTURE WORK

In conclusion, this study has presented an effective way of attacking the information hidden in mail. The method is robust and can successfully recognizes contents of mail enclosed in envelopes of various security patterns. The use of transfer learning in the algorithm has greatly reduced the computational cost and time. Also, the added grammar checking has also improved the results, but there's no significant difference whether or not we have additional information about the surrounding environment to add to the dictionary.

Since the algorithm given in this paper uses context to help understand the rest of the mail, it would be interesting to further study how the amount of context given affects the accuracy of the algorithm. A past study on application-driven data reconstruction analyzed the effect of the percentage of missing data on the model tested and found that the accuracy of the data reconstruction varied greatly depending on the missing data[17]. In this paper, since the algorithm is being applied to the same papers, but with different security patterns, the missing data does not really affect the outcome. For future studies, it would be interesting to analyze the algorithm itself and how greatly the context affects its success.

Some other possible future work in this field include: adding the distance model to the training process, so we can know the how accurate each guess is, thus the grammar module can be more accurate using this additional information. If more data are available, we can also use the additional data to develop a sentence interpreting/recognition mechanism similar to this one.

# REFERENCES

[1] Zhang, Qingquan and Sobelman, Gerald E. and He, Tian. Gradient-based target localization in robotic sensor networks. Pervasive and Mobile Computing, 5:37-48, 2009. [DOI:10.1016/j.pmcj.2008.05.007]

[2] Yi, Ping and Yu, Minjie and Zhou, Ziqiao and Xu, Wei and Zhang, Qingquan and Zhu, Ting. A Three-Dimensional Wireless Indoor Localization System. JECE, 2014. [DOI: 10.1155/2014/149016]

[3] Zhu, Ting and Xiao, Sheng and Zhang, Qingquan and Gu, Yu and Yi, Ping and Li, Yanhua. Emergent Technologies in Big Data Sensing: A Survey. Int. J. Distrib. Sen. Netw., 2015. [DOI:10.1155/2015/902982]

[4] S. Xiao, W. Gong, D. Towsley, Q. Zhang, and T. Zhu. Reliability analysis for cryptographic key management. IEEE International Conference on Communications (ICC), 2013. pp 999-1004.

[5] Zhang, Guoming and Yan, Chen and Ji, Xiaoyu and Zhang, Tianchen and Zhang, Taimin and Xu, Wenyuan. DolphinAttack: Inaudible Voice Commands. Proceedings of the 2017 ACM SIGSAC Conference on Computer and Communications Security. pp 103-117. [DOI:10.1145/3133956.3134052]

[6] Sabourin, Michael and Mitiche, Amar. Optical character recognition by a neural network. Neural Networks, 1992. pp 843-852. [DOI: 10.1016/S0893-6080(05)80144-3]

[7] Krizhevsky, Alex and Sutskever, Ilya and Hinton, Geoffrey E. ImageNet Classification with Deep Convolutional Neural Networks. Proceedings of the 25th International Conference on Neural Information Processing Systems - Volume 1. 2012. pp 1097–1105. [DOI: 10.5555/2999134.2999257]

[8] Zhang, Shiwen and Zhang, Qingquan and Xiao, Sheng and Zhu, Ting and Gu, Yu and Lin, Yaping. Cooperative Data Reduction in Wireless Sensor Network. ACM Trans. Embed. Comput. Syst. 2015. [DOI: 10.1145/2786755]

[9] Vaswani, Ashish and Shazeer, Noam and Parmar, Niki and Uszkoreit, Jakob and Jones, Llion and Gomez, Aidan N. and Kaiser, undefinedukasz and Polosukhin, Illia. Attention is All You Need. Proceedings of the 31st International Conference on Neural Information Processing Systems, 2017. [DOI: 10.5555/3295222.3295349]

[10] Y. Liu and Z. Xia and P. Yi and Y. Yao and T. Xie and W. Wang and T. Zhu. GENPass: A General Deep Learning Model for Password Guessing with PCFG Rules and Adversarial Generation. IEEE International Conference on Communications (ICC), 2018. pp 1-6. [DOI:10.1109/ICC.2018.8422243]

[11] Chi, Zicheng and Yao, Yao and Xie, Tiantian and Liu, Xin and Huang, Zhichuan and Wang, Wei and Zhu, Ting. EAR: Exploiting Uncontrollable Ambient RF Signals in Heterogeneous Networks for Gesture Recognition. Proceedings of the 16th ACM Conference on Embedded Networked Sensor Systems, 2016. pp 237-249. [DOI: 10.1145/3274783.3274847]

[12] Algazi, Allan Stuart (Succasunna, NJ, US). System and methods for mail security. Available from: http://www.freepatentsonline.com/7343299.html. [Last accessed on 18 APR 2020]

[13] Kay, Steven M. Fundamentals of Statistical Signal Processing: Estimation Theory. Prentice-Hall, Inc. USA, 1993. ISBN 0133457117.

[14] Olivas, Emilio Soria and Guerrero, Jose David Martin and Sober, Marcelino Martinez and Benedito, Jose Rafael Magdalena and Lopez, Antonio Jose Serrano. Handbook Of Research On Machine Learning Applications and Trends: Algorithms, Methods and Techniques - 2 Volumes. 2009. [DOI: 10.5555/1803899]

[15] GitHub Repository. Attention-OCR. Available from: https://github.com/da03/Attention OCR/commit/88cff37cad09fd85b5178662130595821d7fe0a2. [Last accessed on: 20 Apr 2020]

[16] Jaderberg M, Simonyan K, Vedaldi A, et al. Synthetic data and artificial neural networks for natural scene text recognition[J]. arXiv preprint arXiv:1406.2227, 2014.

[17] Huang, Z., Xie, T., Zhu, T., Wang, J., & Zhang, Q. Application-driven sensing data reconstruction and selection based on correlation mining and dynamic feedback. In 2016 IEEE International Conference on Big Data (Big Data) (pp. 1322-1327). IEEE. [DOI: 10.1109/BigData.2016.7840737]

[18] Yao, Y., Li, Y., Liu, X., Chi, Z., Wang, W., Xie, T., & Zhu, T. Aegis: An interference-negligible rf sensing shield. In IEEE INFOCOM 2018-IEEE Conference on Computer Communications (pp. 1718-1726). IEEE. [DOI: 10.1109/INFOCOM.2018.8485883]

[19] Li, Y., & Zhu, T. Gait-based wi-fi signatures for privacy-preserving. In Proceedings of the 11th ACM on Asia Conference on Computer and Communications Security (pp. 571-582). [DOI: 10.1145/2897845.2897909]

[20] Li, Y., & Zhu, T. (2016, June). Using wi-fi signals to characterize human gait for identification and activity monitoring. In 2016 IEEE First International Conference on Connected Health: Applications, Systems and Engineering Technologies (CHASE) (pp. 238-247). IEEE. [DOI: 10.1109/CHASE.2016.20]

[21] Huang, Z., Luo, H., & Zhu, T. (2015). Signature-based Detection for Activities of Appliances. In PES. [DOI: 10.1145/2676061.2675045]

[22] Xiao, Sheng, et al. "Reliability analysis for cryptographic key management." 2014 IEEE International Conference on Communications (ICC). IEEE, 2014. [DOI: 10.1109/ICC.2014.6883450]

[23] Yi, P., Zhu, T., Liu, N., Wu, Y., & Li, J. (2012). Cross-layer detection for black hole attack in wireless network. Journal of Computational Information Systems, 8(10), 4101-4109.

[24] Yi P, Zhu T, Ma J, et al. An Intrusion Prevention Mechanism in Mobile Ad Hoc Networks[J]. Ad Hoc & Sensor Wireless Networks, 2013, 17(3-4): 269-292.

[25] Yi P, Zhu T, Zhang Q, et al. A denial of service attack in advanced metering infrastructure network[C]//2014 IEEE International Conference on Communications (ICC). IEEE, 2014: 1029-1034. [DOI: 10.1109/ICC.2014.6883456]

[26] Yi P, Zhu T, Zhang Q, et al. Puppet attack: A denial of service attack in advanced metering infrastructure network[J]. Journal of Network and Computer Applications, 2016, 59: 325-332. [DOI: 10.1016/j.jnca.2015.04.015]

[27] Zhu T, Yu M. A dynamic secure QoS routing protocol for wireless ad hoc networks[C]//2006 IEEE Sarnoff Symposium. IEEE, 2006: 1-4. [DOI: 10.1109/SARNOF.2006.4534795]

[28] Zhu T, Yu M. Nis02-4: A secure quality of service routing protocol for wireless ad hoc networks[C]//IEEE Globecom 2006. IEEE, 2006: 1-6. [DOI: 10.1109/GLOCOM.2006.270]

[29] Chi Z, Li Y, Liu X, et al. Parallel inclusive communication for connecting heterogeneous IoT devices at the edge[C]//Proceedings of the 17th Conference on Embedded Networked Sensor Systems. 2019: 205-218. [DOI: 10.1145/3356250.3360046]

[30] Wang W, Liu X, Yao Y, et al. CRF: coexistent routing and flooding using WiFi packets in heterogeneous IoT networks[C]//IEEE INFOCOM 2019-IEEE Conference on Computer Communications. IEEE, 2019: 19-27. [DOI: 10.1109/INFOCOM.2019.8737525]

[31] Wang W, Xie T, Liu X, et al. Ect: Exploiting cross-technology concurrent transmission for reducing packet delivery delay in iot networks[C]//IEEE INFOCOM 2018-IEEE Conference on Computer Communications. IEEE, 2018: 369-377. [DOI: 10.1109/INFOCOM.2018.8486349]

[32] Chi Z, Huang Z, Yao Y, et al. EMF: Embedding multiple flows of information in existing traffic for concurrent communication among heterogeneous IoT devices[C]//IEEE INFOCOM 2017-IEEE Conference on Computer Communications. IEEE, 2017: 1-9. [DOI:10.1109/INFOCOM.2017.8057109]

[33] Chi Z, Li Y, Yao Y, et al. PMC: Parallel multi-protocol communication to heterogeneous IoT radios within a single WiFi channel[C]//2017 IEEE 25th International Conference on Network Protocols (ICNP). IEEE, 2017: 1-10. [DOI: 10.1109/ICNP.2017.8117550]

[34] Chi Z, Li Y, Sun H, et al. B2w2: N-way concurrent communication for iot devices[C]//Proceedings of the 14th ACM Conference on Embedded Network Sensor Systems CD-ROM. 2016: 245-258. [DOI: 10.1145/2994551.2994561]

[35] Xie T, Huang Z, Chi Z, et al. Minimizing amortized cost of the on-demand irrigation system in smart farms[C]//Proceedings of the 3rd International Workshop on Cyber-Physical Systems for Smart Water Networks. 2017: 43-46. [DOI: 10.1145/3055366.3055370]

[36] Huang Z, Zhu T. Real-time data and energy management in microgrids[C]//2016 IEEE Real-Time Systems Symposium (RTSS). IEEE, 2016: 79-88. [DOI: 10.1109/RTSS.2016.017]

[37] Huang Z, Zhu T. Leveraging multi-granularity energy data for accurate energy demand forecast in smart grids[C]//2016 IEEE International Conference on Big Data (Big Data). IEEE, 2016: 1182-1191. [DOI: 10.1109/BigData.2016.7840722]

[38] Huang Z, Zhu T, Lu H, et al. Accurate power quality monitoring in microgrids[C]//2016 15th ACM/IEEE International Conference on Information Processing in Sensor Networks (IPSN). IEEE, 2016: 1-6. [DOI: 10.1109/IPSN.2016.7460660]

[39] Huang Z, Zhu T, Gu Y, et al. Shepherd: sharing energy for privacy preserving in hybrid AC-DC microgrids[C]//Proceedings of the Seventh International Conference on Future Energy Systems. 2016: 1-10. [DOI: 10.1145/2934328.2934347]

[40] Gao J, Yi P, Chi Z, et al. A smart medical system for dynamic closed-loop blood glucose-insulin control[J]. Smart Health, 2017, 1: 18-33. [DOI: 10.1016/j.smhl.2017.04.001]

[41] Gao J, Yi P, Chi Z, et al. Enhanced wearable medical systems for effective blood glucose control[C]//2016 IEEE First International Conference on Connected Health: Applications, Systems and Engineering Technologies (CHASE). IEEE, 2016: 199-208. [DOI: 10.1109/CHASE.2016.11]

[42] Liu X, Chi Z, Wang W, et al. VMscatter: A Versatile {MIMO} Backscatter[C]//17th {USENIX} Symposium on Networked Systems Design and Implementation ({NSDI} 20). 2020: 895-909.

[43] Xu J, Yi P, Xie T, et al. Charge station placement in electric vehicle energy distribution network[C]//2017 IEEE International Conference on Communications (ICC). IEEE, 2017: 1-6. [DOI: 10.1109/ICC.2017.7996576]

[44] Sui Y, Yi P, Liu X, et al. Optimization for charge station placement in electric vehicles energy network[C]//Proceedings of the Workshop on Smart Internet of Things. 2017: 1-6. [DOI: 10.1145/3132479.3132480]

[45] Huang Z, Xie T, Zhu T, et al. Application-driven sensing data reconstruction and selection based on correlation mining and dynamic feedback[C]//2016 IEEE International Conference on Big Data (Big Data). IEEE, 2016: 1322-1327. [DOI: 10.1109/BigData.2016.7840737]

[46] Huang Z, Luo H, Skoda D, et al. E-sketch: Gathering large-scale energy consumption data based on consumption patterns[C]//2014 IEEE International Conference on Big Data (Big Data). IEEE, 2014: 656-665. [DOI: 10.1109/BigData.2014.7004289]

[47] Tsai, C. W., Lai, C. F., Chao, H. C., & Vasilakos, A. V. (2015). Big data analytics: a survey. Journal of Big data, 2(1), 21. [DOI: 10.1186/s40537-015-0030-3]

[48] Chen, F., Deng, P., Wan, J., Zhang, D., Vasilakos, A. V., & Rong, X. (2015). Data mining for the internet of things: literature review and challenges. International Journal of Distributed Sensor Networks, 11(8), 431047. [DOI: 10.1155/2015/431047]

[49] Dai, H. N., Wong, R. C. W., Wang, H., Zheng, Z., & Vasilakos, A. V. (2019). Big data analytics for large-scale wireless networks: Challenges and opportunities. ACM Computing Surveys (CSUR), 52(5), 1-36. [DOI: 10.1145/3337065]

[50] Gan, W., Lin, J. C. W., Chao, H. C., Vasilakos, A. V., & Philip, S. Y. (2020). Utility-driven data analytics on uncertain data. IEEE Systems Journal. [DOI: 10.1109/JSYST.2020.2979279]

[51] Wan, J., Zheng, P., Si, H., Xiong, N. N., Zhang, W., & Vasilakos, A. V. (2019). An artificial intelligence driven multi-feature extraction scheme for big data detection. IEEE Access, 7, 80122-80132. [DOI: 10.1109/ACCESS.2019.2923583]

[52] Kumar, N., Vasilakos, A. V., & Rodrigues, J. J. (2017). A multi-tenant cloud-based DC nano grid for self-sustained smart buildings in smart cities. IEEE Communications Magazine, 55(3), 14-21. [DOI: 10.1109/MCOM.2017.1600228CM]

[53] Yan, Z., Zhang, P., & Vasilakos, A. V. (2016). A security and trust framework for virtualized networks and software‐defined networking. Security and communication networks, 9(16), 3059-3069. [DOI:10.1002/sec.1243]

[54] Mohammad Wazid, et al. LAM-CIoT: Lightweight authentication mechanism in cloud-based IoT environment. J. Netw. Comput. Appl. 150 (2020) [DOI: 10.1016/j.jnca.2019.102496]

[55] Zhou, J., Cao, Z., Dong, X., & Vasilakos, A. V. (2017). Security and privacy for cloud-based IoT: Challenges. IEEE Communications Magazine, 55(1), 26-33. [DOI: 10.1109/MCOM.2017.1600363CM]

[56] Jing, Q., Vasilakos, A. V., Wan, J., Lu, J., & Qiu, D. (2014). Security of the Internet of Things: perspectives and challenges. Wireless Networks, 20(8), 2481-2501. [DOI: 10.1007/s11276-014-0761-7]

[57] Lin, C., He, D., Huang, X., Choo, K. K. R., & Vasilakos, A. V. (2018). BSeIn: A blockchain-based secure mutual authentication with fine-grained access control system for industry 4.0. Journal of Network and Computer Applications, 116, 42-52. [DOI: 10.1016/j.jnca.2018.05.005]

[58] Wazid, M., Das, A. K., Kumar, N., & Vasilakos, A. V. (2019). Design of secure key management and user authentication scheme for fog computing services. Future Generation Computer Systems, 91, 475-492. [DOI: 10.1016/j.future.2018.09.017]

[59] Yu, Y., Xue, L., Au, M. H., Susilo, W., Ni, J., Zhang, Y., ... & Shen, J. (2016). Cloud data integrity checking with an identity-based auditing mechanism from RSA. Future Generation Computer Systems, 62, 85-91. [DOI: 10.1016/j.future.2016.02.003]

[60] Wazid, M., Das, A. K., Kumar, N., Vasilakos, A. V., & Rodrigues, J. J. (2018). Design and analysis of secure lightweight remote user authentication and key agreement scheme in Internet of drones deployment. IEEE Internet of Things Journal, 6(2), 3572-3584. [DOI: 10.1109/JIOT.2018.2888821]

[61] Chen, J., Zhou, J., Cao, Z., Vasilakos, A. V., Dong, X., & Choo, K. K. R. (2019). Lightweight Privacy-preserving Training and Evaluation for Discretized Neural Networks. IEEE Internet of Things Journal. [DOI: 10.1109/JIOT.2019.2942165]

[62] Sun, G., Xu, Z., Yu, H., Chen, X., Chang, V., & Vasilakos, A. V. (2019). Low-latency and resource-efficient service function chaining orchestration in network function virtualization. IEEE Internet of Things Journal. [DOI: 10.1109/JIOT.2019.2937110]

[63] Sun, G., Zhou, R., Sun, J., Yu, H., & Vasilakos, A. V. (2020). Energy-Efficient Provisioning for Service Function Chains to Support Delay-Sensitive Applications in Network Function Virtualization. IEEE Internet of Things Journal. [DOI: 10.1109/JIOT.2020.2970995]

[64] Huang, M., Liu, A., Xiong, N. N., Wang, T., & Vasilakos, A. V. (2020). An Effective Service-Oriented Networking Management Architecture for 5G-Enabled Internet of Things. Computer Networks, 107208. [DOI: 10.1016/j.comnet.2020.107208]

[65] Pan, Y., Li, S., Chang, L. J., Yan, Y., & Zhu, T. (2019). An Unmanned Aerial Vehicle Navigation Mechanism with Preserving Privacy. IEEE International Conference on Communications. [DOI: 10.1109/ICC.2019.8761533]

[66] Pan, Y., Bhargava, B., Ning, Z., Slavov, N., Liu, J., Xu, S., Li, C., & Zhu, T. (2019). Safe and Efficient UAV Navigation Near an Airport. IEEE International Conference on Communications. [DOI: 10.1109/ICC.2019.8761375]

[67] Yi, P., Zhu, T., Zhang, Q., Wu, Y., Li, J. (2012). Green Firewall: an Energy-Efficient Intrusion Prevention Mechanism in Wireless Sensor Network. IEEE Global Communication Conference. [DOI: 10.1109/GLOCOM.2012.6503580]

[68] Zhu, T., Towsley, D. (2011). E$^2$R: Energy Efficient Routing for Multi-hop Green Wireless Network. IEEE INFOCOM Workshop on Green Communications and Networking. [DOI: 10.1109/INFCOMW.2011.5928821]

[69] Jun, J., Gu, Y., Cheng, L., Lu, B., Sun, J., Zhu, T., Niu, J. (2013). SocialLoc: Improving Indoor Localization with Social Sensing. ACM Conference on Embedded Networked Sensor Systems. [DOI: 10.1145/2517351.2517352]

[70] Pei, Z., and Zhu, T. (2011). A Multi-information Localization Algorithm in Wireless Sensor Networks and its Application in Medical Monitoring Systems. ACM Conference on Embedded Networked Sensor Systems. [DOI: 10.1145/2426656]

[71] Huang, Z., Zhu, T., Gu, Y., Irwin, D., Mishra, A., & Shenoy, P. (2014). Minimizing Electricity Costs by Sharing Energy in Sustainable Microgrids. ACM Conference on Embedded Systems for Energy-Efficient Buildings. [DOI: 10.1145/2674061.2674063]

[72] Yi, P., Tang, Y., Hong, Y., Shen, Y., Zhu, T., Zhang, Q., Begovic, M. M. (2014). Renewable Energy Transmission through Multiple Routes in a Mobile Electrical Grid. IEEE Innovative Smart Grid Technologies Conference. [DOI: 10.1109/ISGT.2014.6816468]

[73] Yi, P., Zhu, T., Lin, G., Jiang, X., Li, G., Si, L., Begovic, M. M. (2013). Energy Scheduling and Allocation in Electric Vehicle Energy Distribution Networks. IEEE Power & Energy Society Innovative Smart Grid Technologies Conference. [DOI: 10.1109/ISGT.2013.6497886]

[74] Arikan, A., Jin, R., Wang, B., Han, S., Pattipati, R. K., Yi, P., Zhu, T. (2015). Optimal Renewable Energy Transfer via Electrical Vehicles. IEEE Power & Energy Society Innovative Smart Grid Technologies Conference. [DOI: 10.1109/ISGT.2015.7131839]

[75] Mishra, A., Irwin, D., Shenoy, P., Shen, Y., Kurose, J., and Zhu, T. (2013). GreenCharge: Managing Renewable Energy in Smart Buildings. IEEE Journal on Selected Areas in Communications: Smart Grid Communications Series. [DOI: 10.1109/JSAC.2013.130711]

[76] Zhou, Z., Xie, M., Zhu, T., Xu, W., Yi, P., Huang, Z., Zhang, Q., and Xiao, S. (2014). EEP2P: An Energy-Efficient and Economy-Efficient P2P Network Protocol. IEEE International Green Computing Conference. [DOI: 10.1109/IGCC.2014.7039171]

[77] Zhong, W., Huang, Z., Zhu, T., Gu, Y., Zhang, Q., Yi, P., Jiang, D., and Xiao, S. (2014). iDES: Incentive-Driven Distributed Energy Sharing in Sustainable Microgrids. IEEE International Green Computing Conference. [DOI: 10.1109/IGCC.2014.7039166]

[78] Chi, Z., Yao, Y., Xie, T., Huang, Z., Hammond, M., & Zhu, T. (2016). Harmony: Exploiting coarse-grained received signal strength from IoT devices for human activity recognition. IEEE 24th International Conference on Network Protocols (ICNP) (pp. 1-10). IEEE. [DOI: 10.1109/ICNP.2016.7784414]

[79] Zhu, T., Xiao, S., Ping, Y., Towsley, D., and Gong, W. (2011). A secure energy routing mechanism for sharing renewable energy in smart microgrid. IEEE International Conference on Smart Grid Communications (pp. 143-148). DOI: 10.1109/SmartGridComm.2011.6102307]

[80] Jiang, D., Yuan, Z., Zhang, P., Miao, L., and Zhu, T. (2016). A traffic anomaly detection in communication networks for applications of multimedia medical devices. In Multimedia Tools and Applications (pp. 1-25). [DOI: 10.1007/s11042-016-3402-6]